\documentclass[lettersize,journal]{IEEEtran}
\usepackage{amsmath,amsfonts}
\usepackage{algorithm,algpseudocode}
\usepackage{array}
\usepackage[caption=false,font=normalsize,labelfont=sf,textfont=sf]{subfig}
\usepackage{textcomp}
\usepackage{stfloats}
\usepackage{url}
\usepackage{verbatim}
\usepackage{graphicx}
\usepackage{cite}
\usepackage{soul}

\usepackage{algorithm}
\usepackage{xcolor,colortbl}

\begin{document}

\title{Neuromorphic Auditory Perception by Neural Spiketrum}

\author{
  Huajin Tang,
Pengjie Gu,
Jayawan Wijekoon,
MHD Anas Alsakkal,
Ziming Wang,
Jiangrong Shen,
and Rui Yan
\thanks{This work was supported by National Key Research and Development Program of
        China under Grant No. 2020AAA0105900 and
        National Natural Science Foundation of China under Grant No. 62236007
        (Corresponding author: Huajin Tang.)}
\thanks{Huajin Tang is with the State Key Lab of Brain-Machine Intelligence, College of Computer Science, Zhejiang University, Hangzhou 310027, China (e-mail: {htang@zju.edu.cn}).}
\thanks{Pengjie Gu, Ziming Wang and Jiangrong Shen are with the College of Computer Science, Zhejiang University, Hangzhou 310027, China.}
\thanks{Jayawan Wijekoon and MHD Anas Alsakkal are with the Department of Electrical \& Electronic Engineering, University of Manchester, UK.}
 \thanks{Rui Yan is with the College of Computer Science, Zhejiang University of Technology, Hangzhou 310014, China.}

}

\markboth{Journal of \LaTeX\ Class Files,~Vol.~14, No.~8, August~2021}%
{Shell \MakeLowercase{\textit{et al.}}: A Sample Article Using IEEEtran.cls for IEEE Journals}


\maketitle

\begin{abstract}
  Neuromorphic computing holds the promise to achieve the energy efficiency and robust learning performance of biological neural systems. To realize the promised brain-like intelligence, it needs to solve the challenges of the neuromorphic hardware architecture design of biological neural substrate and the hardware amicable algorithms with spike-based encoding and learning. Here we introduce a neural spike coding model termed spiketrum, to characterize and transform the time-varying analog signals, typically auditory signals, into computationally efficient spatiotemporal spike patterns. It minimizes the information loss occurring at the analog-to-spike transformation and possesses informational robustness to neural fluctuations and spike losses. The model provides a sparse and efficient coding scheme with precisely controllable spike rate that facilitates training of spiking neural networks in various auditory perception tasks. We further investigate the algorithm-hardware co-designs through a neuromorphic cochlear prototype which demonstrates that our approach can provide a systematic solution for spike-based artificial intelligence by fully exploiting its advantages with spike-based computation.
\end{abstract}


\section{Introduction}

\IEEEPARstart{S}pike-based computing paradigm holds the promise of achieving brain-like intelligence and energy efficiency, leveraging on spiking neural networks (SNNs) running on dedicated neuromorphic hardware platforms\cite{pei2019towards,merolla2014million,furber2014spinnaker,schemmel2010wafer}. The synergistic hardware-algorithm paradigm, aka, neuromorphic computing, requires the close emulation of computational principles of the brain, including co-location of memory and computing\cite{Giacomo2015,Abbott2000Synaptic}, sparse spatiotemporal spikes\cite{song2000competitive,Maass2015,Mem2014}, and complex neural-synaptic dynamics\cite{Lillicrap2016Random,G2016Spiking,Nicola2017Supervised}. Increasing developments of spike-based hardware\cite{benjamin2014neurogrid,davies2018loihi,Liu2016,Qiao2015} aim to improve intelligent processing capabilities of day-to-day used interactive devices, adopting amazingly complex, intelligent processing techniques evolved in biological systems. Moreover, the technological progressions of low-power, fast, intelligent decision making neuromorphic systems\cite{Tenore2011Neuromorphic} demand the need to embed rich attentive signal features and wide-ranging background information contained in the surroundings using an efficient spike-based representation. However, these developments of spike-based learning and hardware architecture design\cite{cao2015spiking,Shrestha2018,Emre2019,roy2019towards} rely on simple arithmetic conversion methods that convert a static value to a spike train with a fixed frequency, losing not only precision and but also adaptability to sound structures and environmental noises.

For auditory perception tasks, in contrast to the simple arithmetic coding methods employed by the current neuromorphic framework, our brains solve these tasks efficiently in a very distinct way by neural coding\cite{Ding2012Neural,Schneider2013Sparse,Oxenham2018How}. Throughout the auditory pathway, the time-relative structures and spectral acoustic signatures of input sound waveforms will be represented by the spatiotemporal spike patterns\cite{Rhode1994Encoding,Krishna2000Auditory,Bartlett2007Neural} in which the hidden and high-level auditory features can be detected and learned by downstream spiking neurons\cite{Sharpee2011Hierarchical}.
The widely adopted neural coding algorithms\cite{Dennis2015Combining,Sompolinsky2009Time,Schafer2014Noise} aim to extract particular acoustic features from the auditory image and map them into spike patterns, such as the local spectrogram peaks\cite{Dennis2015Combining}, the onsets or offsets on the intensity of spectrogram in a given frequency band\cite{Sompolinsky2009Time,singh2003modulation} and the responses of an artificial neuron with an adaptable spatiotemporal receptive field (STRF)\cite{Schafer2014Noise,Tavanaei2016A}.
Nevertheless, these techniques are infeasible to reveal the real acoustic structures and to capture the high-level auditory representations which could not get promised performance in precisely controllable spike rate.
Furthermore, the overlapping and block-based representations of auditory images may mostly obscure or distort the transients and nonstationary structures of sounds.
These precise temporal structures are crucial for auditory perception\cite{Wang2008Neural,Elhilali2004Dynamics,Schnupp2006Plasticity,Araki2016Mind}, as the time-varying sound signals are temporally modulated by the brain across a wide range of time scales from a few milliseconds to tens and hundreds of milliseconds\cite{Rosen1992Temporal}. The slow-varying temporal modulations are important for speech recognition\cite{holmberg2007speech,Schnupp2006Plasticity,Rosen1992Temporal} and melody perception\cite{Darwin2005Pitch}, while the rapid-changing modulations yield other types of sensations like pitch\cite{Oxenham2018How,Darwin2005Pitch}, sound localization\cite{Furukawa2002Cortical}, and sound roughness\cite{Rosen1992Temporal,Wang2008Neural}.
The information loss and distortion on the temporal representation would adversely affect the performance of spike-based learning.
In contrast, some cochlea-like models\cite{zilany2014updated,holmberg2007speech,Rudnicki2015Modeling}, from a bottom-up perspective, attempt to mimic all relevant physiological processes, such as the function of inner hair cells and the ribbon synaptic mechanism, over the whole length of the hearing organ and to integrate them into one model.
While mimicking the detailed dynamics of complex biological sensory architectures more closely contribute to an understanding of techniques used by the biological apparatus, these models request higher computational complexity and are not available to control the spike rate to optimize the available computational resource\cite{Rudnicki2015Modeling}, and moreover, direct hardware implementation of those techniques may not always translate into efficient, low-power implementations\cite{leong2003fpga,Liu2010Neuromorphic,liu2014event}.

We here introduce a neural spike coding model termed as {\it spiketrum}, encoding auditory signals by sparse spatiotemporal spike trains.
{\it Spiketrum} represents sounds with a minimum of resources by adapting to the statistical structure with precisely controllable spike rate. It is fully compatible to neuromorphic techniques based on spiking neurons and hardware platforms. A neuromorphic cochlea prototype based on the spiketrum model demonstrates its ability to represent rich audio signals efficiently using spike trains in real-time, making it an ideal candidate as a front-end for neuromorphic devices and spike-based intelligent computing machines, such as autonomous robots and cognitive computers.

\begin{figure*}[htb!] 
	\centering
$\begin{array}{cc}
\includegraphics[width= 0.7\linewidth]{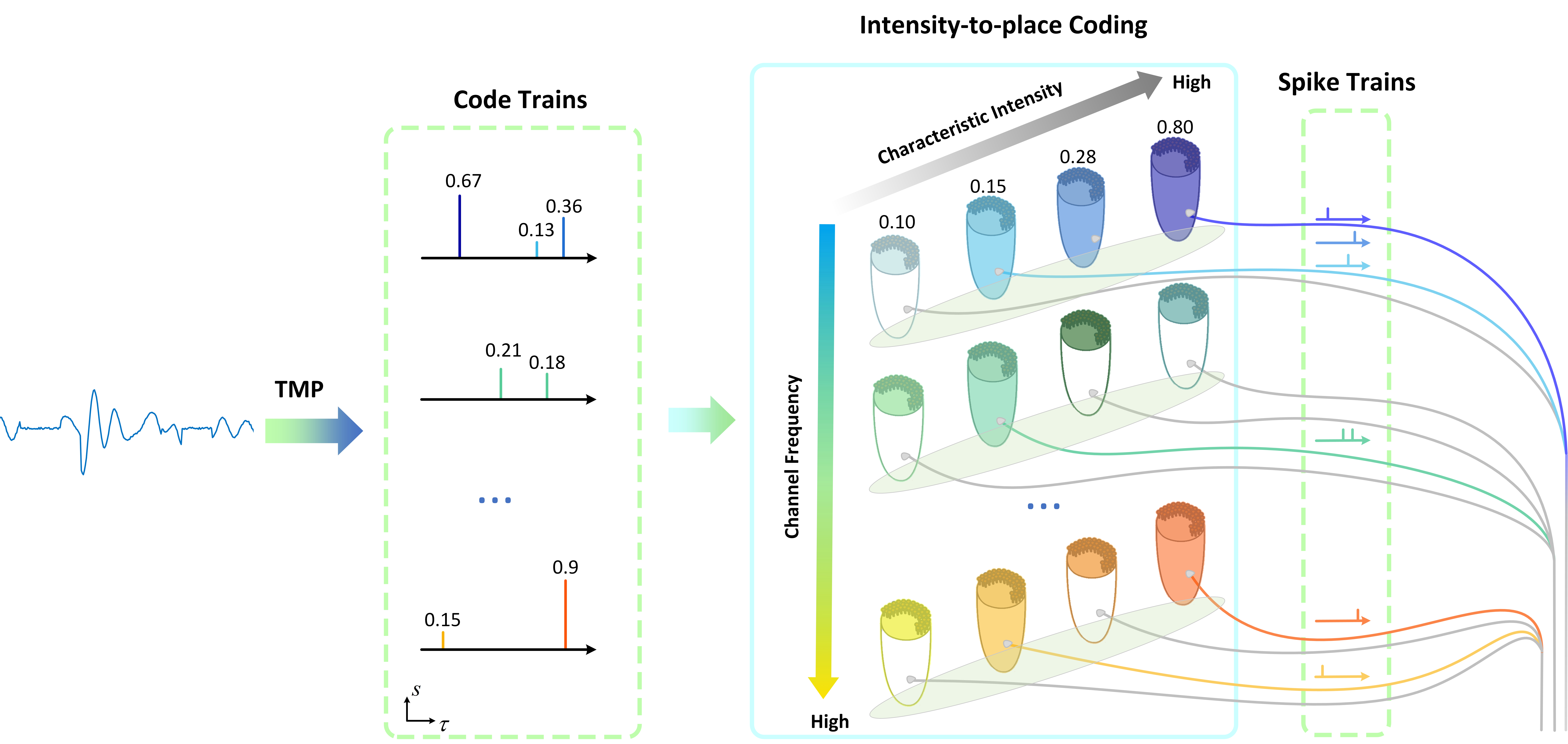}\\
	 \includegraphics[width= 0.7\linewidth]{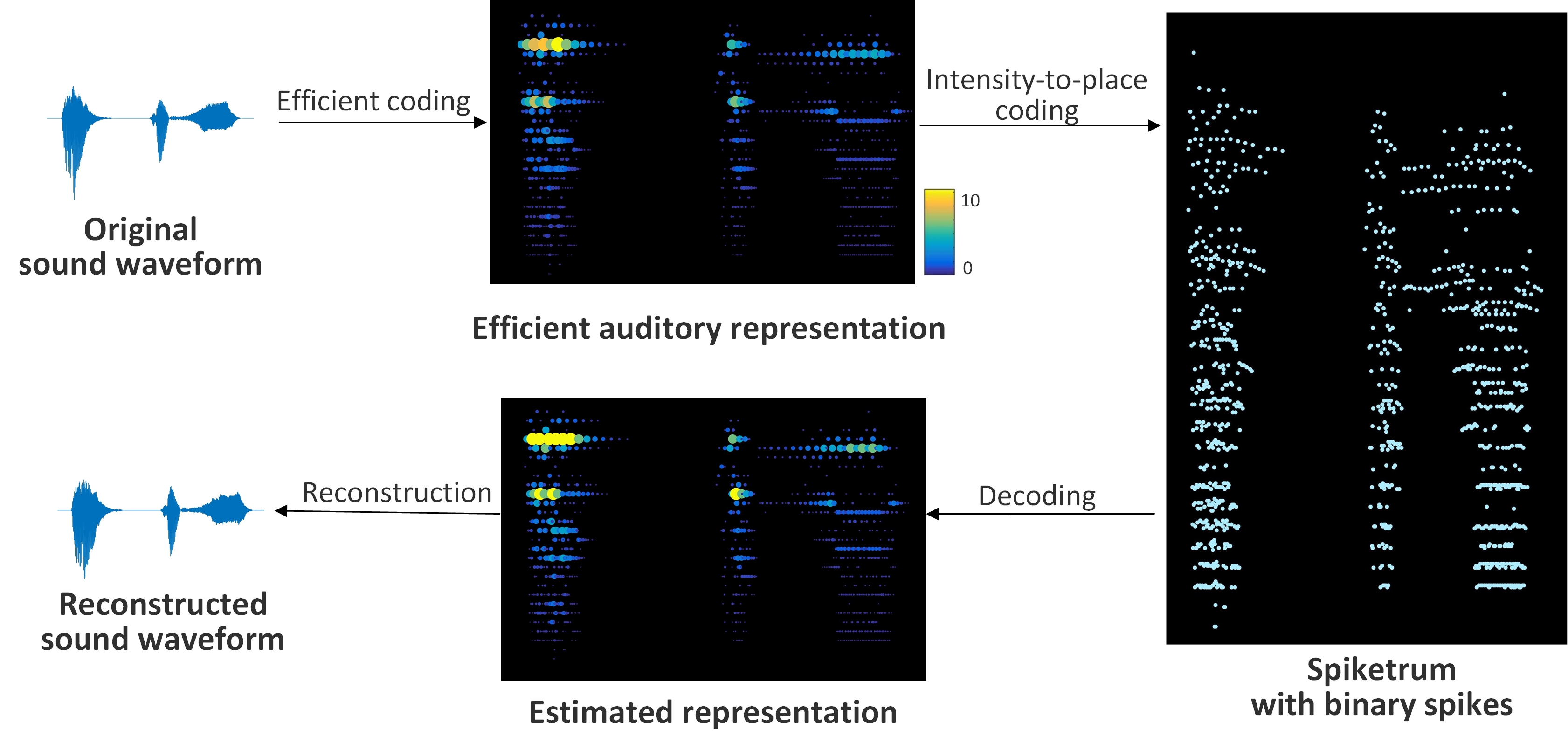}
\end{array}$
	\caption{Schematic of auditory coding by the neural spiketrum. An original sound waveform will be encoded into an efficient auditory representation by utilizing an efficient coding theory. The proposed intensity-to-place coding will convert this representation into binary neural spikes by reflecting the intensity information through neural spatial information. Importantly, we can decode the spiketrum to estimate the efficient auditory representation and reconstruct the original sound waveform.
	}
\label{fig.model}
\end{figure*}

\section{ The \emph{Spiketrum} auditory encoding model}

We aim to develop an auditory encoding model termed as \emph{spiketrum} that could maximize the amount of perceptual information carried in neural representations while minimizing the cost for transmission and information loss (Fig. \ref{fig.model}).
Adhering to this fundamental principle, \emph{spiketrum} presents a comprehensive resolution for the sparse decomposition and binarization of time-varying signals with biologically feasible kernels. We elaborate on the primary constituents of the neural spiketrum model from three key facets: efficient event-based coding, intensity-to-place coding and kernel set selection.


\subsection{Efficient Even-Based Coding}
\label{sc.eec}
In order to encode sound signals into sparse codes with precise temporal information, a sound waveform $x(t)$ can be regarded as a linear superposition of a set of sparse, time-shifted kernels $\phi_{m_i}\in \Phi$, $m_i=1,\cdots,M$  based on the efficient coding theory\cite{Smith2006Efficient}. Each kernel can be positioned arbitrarily and independently in time. The mathematical description of the representation is as follows:
\begin{equation}
x(t)=\sum_{i} s_i \phi_{m_i} (t-\tau_i) + \epsilon(t)
\label{eq.sparse_coding}
\end{equation}
where $\tau_i$ and $s_i$ are the temporal position and amplitude (also called intensity) of the $i$-th instance $\phi_{m_i}(t)$, $i=1,\cdots,N$, and the additive noise $\epsilon(t)$ denotes the residual between the input signal $x(t)$ and the reconstruction $\widehat{x}(t)$. The number of kernels $M$, the length and shape of each kernel $\phi_{m_i}$ are purely arbitrary, which need to determine before encoding an acoustic signal.

In the next, we will require an algorithm to find the optimal sparse codes $(m_i,\tau_i,s_i)$ of acoustic signals that minimizes the residual $\epsilon(t)$ according to Eq. \ref{eq.sparse_coding}. While reconstructing the original signal from codes is a linear operation, deriving the optimal codes for a signal is highly nonlinear and computationally complex\cite{Davis1997Adaptive}. Consequently, there are various encoding algorithms to get different approximate solutions of the efficient coding model. Here, we adopt primarily the temporal matching pursuit (TMP) algorithm to get codes for temporal signals and propose the Event-based Temporal Matching Pursuit (E-TMP) algorithm. Among all possible solutions, the solutions of matching pursuit algorithm are absolutely sparse and highly efficient. Only those most correlated spike codes, which possess the maximal amplitude coefficients, are extracted to represent signals. Specifically, the algorithm is an iterative method which finds one most correlated code and remove the influence of the code from the original signal in each iteration. So the number of spike codes is controllable in this algorithm\cite{Krstulovic2011Mptk}, which is equal to the iteration number $N$ of E-TMP algorithm, and the spike rate $\lambda$ of \emph{spiketrum} is formulated as
 \begin{equation}
 \lambda = N/\tau
 \label{eq.spikeate}
 \end{equation}
 where $\tau$ is the signal duration. It yields $N$ spikes with fine temporal resolution in contrast to spectrogram-based method.
 The detailed procedure of E-TMP algorithm is shown in Algorithm \ref{alg::MatchingPursuit}.

 \begin{algorithm}[t]
  \caption{Event-based Temporal Matching Pursuit (E-TMP) Algorithm}
  \label{alg::MatchingPursuit}
  \begin{algorithmic}[1]
    \Require

    Sound signal: $x(t)$

    Kernel set: $\Phi = \{\phi_1(t)\ldots\phi_M(t)\}$

    The maximum number of iterations: $N$

    The minimum energy ratio: $\epsilon_{min}$
    \Ensure
    Sound spike codes:

    $S = \{(m_1,\tau_1,s_1)\ldots(m_N,\tau_N,s_N)\}$
    \State Initialize: $R_1 \gets x(t)$, $i \gets 0$, $S \gets \{\}$
    \Repeat
    \State $i \gets i+1$
    \State Compute the cross-correlations of $R_i$ and each kernel $\phi_m$ in $\Phi$, using sliding inner products:
    \begin{equation*}
    H_i^m (\tau) =\int_{-\infty}^{+\infty}R_i(t)\phi_m(t+\tau)dt
    \end{equation*}

    \State Search for the most correlated kernel and its most activated temporal position by finding the maximum value of the correlations:
    \begin{equation*}
    (m_i,\tau_i)=\mathop{\arg \max}_{m,\tau} \ \ H_i^m(\tau)
    \end{equation*}
    \State Remove the influence of kernel $\phi_{m_i}$ at temporal position $\tau_i$ from $R_i$:
    \begin{equation*}
    R_{i+1}(t) = R_{i}(t) - H_i^{m_i} (\tau_i)\phi_{m_i}(t-\tau_i)
    \end{equation*}
    \State Add the code to $S$:
    \begin{equation*}
    S = S \bigcup \{(m_i,\tau_i,s_i)\}
    \end{equation*}
    where  $s_i =H_i^{m_i}(\tau_i)$
    \Until{($n>N \quad or \quad \frac{{\|R_i\|}^2}{{\|R_1\|}^2}< \epsilon_{min}$)}

  \end{algorithmic}
\end{algorithm}

 \subsection{Intensity-To-Place Coding}
After generating the sparse events through E-TMP algorithm,
the analog intensities need to be converted into all-or-none spikes that spiking neurons
can process\cite{roy2019towards,Time2004}. Here we propose a deterministic approach named intensity-to-place (ITP) coding: the energy distribution of sounds can be reflected through the spatial positions of neurons.

As the precise temporal structures are crucial acoustic clues for auditory tasks, we first normalize all scaling coefficients in efficient coding process into $[0,1]$, without changing the temporal positions of spikes.
All codes $(m_i,\tau_i,s_i)$ with the same kernel $\phi_{m_i}$ are regarded as a code train, and each code train is stretched into $K$ neurons. Assuming that each neuron has a characteristic intensity $c_k$ obeying a logarithmic normal distribution, each code in a code train is then able to be deterministically encoded by one specific neuron with the minimum intensity distance, firing a spike at $\tau_i$ and mapping into the specified neuron position.
Formally, let $c_1,\cdots,c_K$ denote the characteristic intensities of $K$ neurons, then each code train will stretch out into $K$ neurons.
In particular, a code $(m_i,\tau_i,s_i)$ will be routed into the $h$-th neuron according to its coefficient $s_i$ and converted into a spike generated at the same temporal position $\tau_i$.
$h$ is the neuron position in the spatiotempotral spike pattern given by
\begin{eqnarray}
k &=& \mathop{\arg \min}_k \ \ \lvert c_k - s_i \rvert, \\
h &=& K(m_i-1)+k
\end{eqnarray}
where $i=1,\cdots,N$, $ m_i=1,\cdots,M$.

The above process is possibly supported by neural circuits performing winner-take-all (WTA) computation widely found in the brain\cite{flyScience}, which has the advantage
of being
fully implemented by spiking neural network using VLSI circuits\cite{oster2004winner}. The other advantage lies in that neural population coding could extend the range of sound levels that can be accurately encoded, fine-tuning hearing to the local acoustic environment\cite{dean2005neural}.

It is crucial to select an appropriate set of center intensities for $c_k$, which has a direct effect on the representational precision and information organization of spiketrums. The simplest selection strategy is linear. It only needs to select centers which are equally spaced from zero to a given value. However, code intensities coeffcients of natural sounds follow an exponential distribution in which most intensity coeffificients values are close to zero. In this way, linear strategy would generally assign the binary spikes to the synapses whose corresponding center intensity value are close to zero, while other synapses with higher intensities would generate few spikes. Apparently, it is difficult for linear strategy to utilize place information of spiketrums efficiently, since it brackets together most code intensity coefficients and thus reduces the representational precision of spiketrums. To improve representational precision, we have designed another selection strategy which tends to assign spikes to synapses with different corresponding intensities. This strategy would select center intensities whose logarithmic values are equally spaced from zero to a given value, because we found that logarithmic of intensity values are subject to an approximation of Gaussian distribution.

 \noindent
 \subsection{ Gammatone Kernel Set}
 The kernel set $\Phi$ for $\phi_{m_i}\in \Phi$ should be determined before encoding, and while the length and shape of each kernel can be purely arbitrary, the coding efficiency and representation precision of \emph{spiketrum} depends crucially on $\Phi$.
 In general, the great ability of kernels to capture the inherent acoustic structures of the sound ensemble can contribute significantly to accurate representations of sounds with a limited number of spikes.

 There is growing consensus that Gammatone functions can simulate the modulated properties of auditory nerve fibers. Therefore, we select a group of equivalent rectangular band (ERB) Gammatone filters as the kernel set of the efficient coding model, expecting to get a biologically plausible coding
 that might show qualitative similarities to the neural spike coding generated at mammalian cochleas\cite{Lewicki2002Efficient}. The characteristic frequencies of these filters are distributed logarithmically and follow the frequency resolution observed in
 biological
 cochlear filters --- a large portion of filters are centered around low frequencies\cite{Valero2012Gammatone}.
  Consequently, we believe Gammatone kernels can capture the underlying structure of a broad range of sounds. A Gammatone filter is a linear filter described by an impulse response that is the product of a gamma distribution and sinusoidal tone. It is a widely used model of auditory filters in the auditory system. The Gammatone impulse response is given by:
 \begin{equation}
 g(t) = at^{n-1}e^{-2\pi bt} \cos(2\pi ft +\phi)
 \label{eq.gammatone}
 \end{equation}
 where $f$ (in Hz) is the characteristic frequency, $\phi$ (in radian) is the phase of the carrier, $a$ is the amplitude, $n$ is the filter's order, $b$ (in Hz) is the filter's bandwidth, and $t$ (in seconds) is time.
 The higher the frequency of a Gammatone function, the denser its waveform and the shorter its length.

\section{Performance Analysis of Neural Spiketrum}
For various experimental purposes, we utilize diverse sound samples from multiple sound datasets to demonstrate that Spiketrum presents a sparse, precise, and robust spike representation for continuous waveforms. These include a corpus of natural sounds, sound events, solo music, and speech. Further details regarding the utilized data are provided in the Supplementary Material.

\subsection{ The Sparse, Precise and Robust Representation of Spiketrum}

We will first demonstrate that the spiketrum model can encode the underlying time-relative structure and the essential acoustic attributes (e.g., frequency and sound energy) of auditory information precisely by the spatiotemporal spike patterns (Fig.\ref{fig.model}, right), and then we investigate the robustness to spike fluctuations and signal noises of spiketrum.

Motivated by the fact that natural sounds with most of their modulation energy for low temporal and spectral modulations are low-passed in general\cite{singh2003modulation}, i.e., intensity coefficients of natural sounds follow a logarithmic normal distribution, we hypothesize that the characteristic intensities $c_k$ of encoding neurons follow a same distribution in ITP. To verify the efficacy of the hypothesis, we choose linear distribution as the baseline for comparison, and analyze the representation precision by using a collective of sound samples\cite{Normanhaignere2015Distinct} for the two different intensity distributions, denoted by $P_{log}$ and $P_{linear}$ respectively. The results show that ITP enables the generated binary spikes to spread over more neurons, implying that there are more active neurons jointly representing the auditory information, more efficient than taking linear distribution which generates most spikes on neurons of small characteristic intensities.

\begin{figure}[t]
	\centering
	\includegraphics[width=\linewidth]{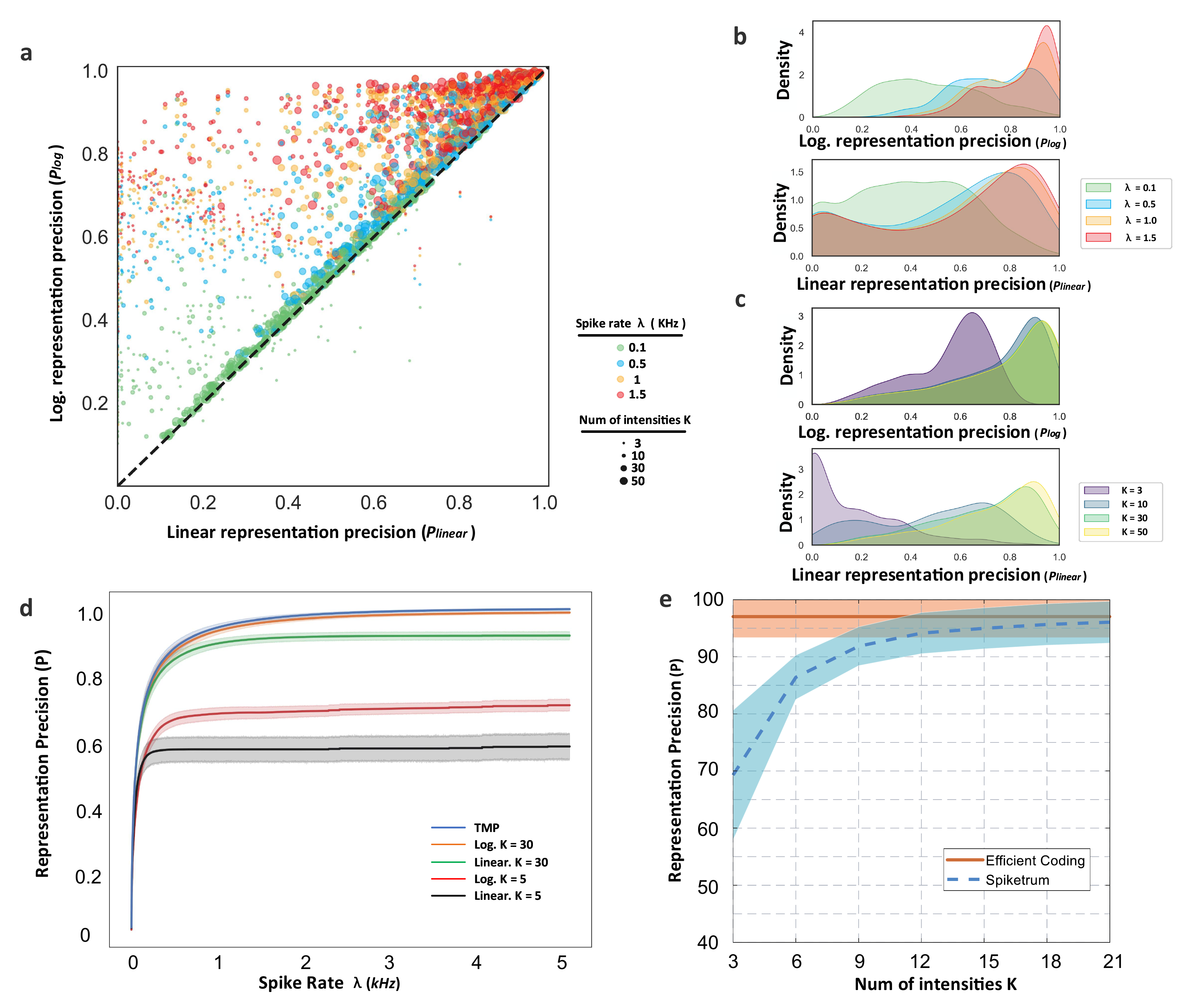}
	\caption{ The representational precision $P$ of \emph{spiketrum} under different coding parameters.
		\textbf{a} The scatter plot of $P_{linear}$ and $P_{log}$ for a variation of $\lambda$ and $K$. For 165 sound samples, the color and size denoting $\lambda$ and $K$ respectively, there are 2640 dots in total. Top and right subplots show the marginal distributions of $P_{linear}$ and $P_{log}$ under different spike rate.
		\textbf{b,c} The curves of $P$ and averages over all 165 sound samples for different coding schemes. 
			\textbf{d} With the spike rate
		increasing from 0 to 5kHz (at an interval of 1Hz), $P$ continues to increase until saturation. Selecting as few as 30 center intensities ($K=30$), $P_{log}$ is close to that of E-TMP algorithm.
		\textbf{e} The reconstruction accuracy in analog spikes (yellow) and $spiketrum$ (blue) with different rank (size of different center intensity group). As the rank increases, the reconstruction accuracy of $spiketrum$  gradually increases and approaches that of  the efficient auditory representation. The experiments were run on nearly 1000 random samples from RWCP dataset. }
	\label{fig.strategies_acc}
\end{figure}

 The representation precision analysis shows that higher values of spike rate $\lambda$ and number of characteristic intensities $K$ both lead to higher precision, and in particular, the representation precision with $P_{log}$ of all sound samples are higher than that with $P_{linear}$ (Fig. \ref{fig.strategies_acc}a).
When $\lambda$ is high enough, its increase no longer promotes increases of $P_{log}$ and $P_{linear}$ obviously (Fig. \ref{fig.strategies_acc}b, d).
For small number of $K$, e.g., $K=5$, $P_{log}$ is mainly distributed around 0.6, significantly better than $P_{linear}$ which is gathered the left of $0.4$ (Fig. \ref{fig.strategies_acc}c), confirming the efficiency of the hypothesis for ITP.
It is worth noting
that $P_{log}$ approaches closely with the values computed by the efficient coding theory (Fig. \ref{fig.strategies_acc}d, e) even setting $K$ as small as $K=30$.
The property that the coding efficiency of \emph{spiketrum} can be well predicted by $K$ as well as $\lambda$ which is precisely controllable (see Section \ref{sc.eec}), turns out to be an appealing feature of \emph{spiketrum} as it can be customized to optimize the constrained biological resources or neuromorphic hardware.

\noindent
\subsection{ Robustness to Spike Fluctuations and Noises}
Neural population coding can stave off the catastrophic consequences of the fluctuation of neural responses or damage to single neurons\cite{Jazayeri2006Optimal,Urbanczik2009Reinforcement}; also, biological coding in primary auditory cortex is stable to behaviorally relevant sounds despite a variety of distracting environmental noises\cite{Schneider2013Sparse,mesgarani2014mechanisms}. With above considerations in mind, we expect to check whether \emph{spiketrum} is robust to environmental noises and spike fluctuations.
Here, we analyze the robustness of \emph{spiketrum} from the perspective of the information theory\cite{Shannon1948A}, in which an analytical framework for quantitative comparing of the sensory signals and spike patterns is available. Meanwhile, nonlinear dependencies of any statistical order\cite{Kayser2009Spike} can be captured. Unfortunately, it is computationally complex and data-hungry to directly compute the amount of information conveyed by spike patterns. Therefore, we employ an alternative Fourier method to approximate the information entropy of spike patterns\cite{Crumiller2013The} (see Supplementary Materials).
\begin{figure}[t] 
	\centering
	\includegraphics[width= \linewidth]{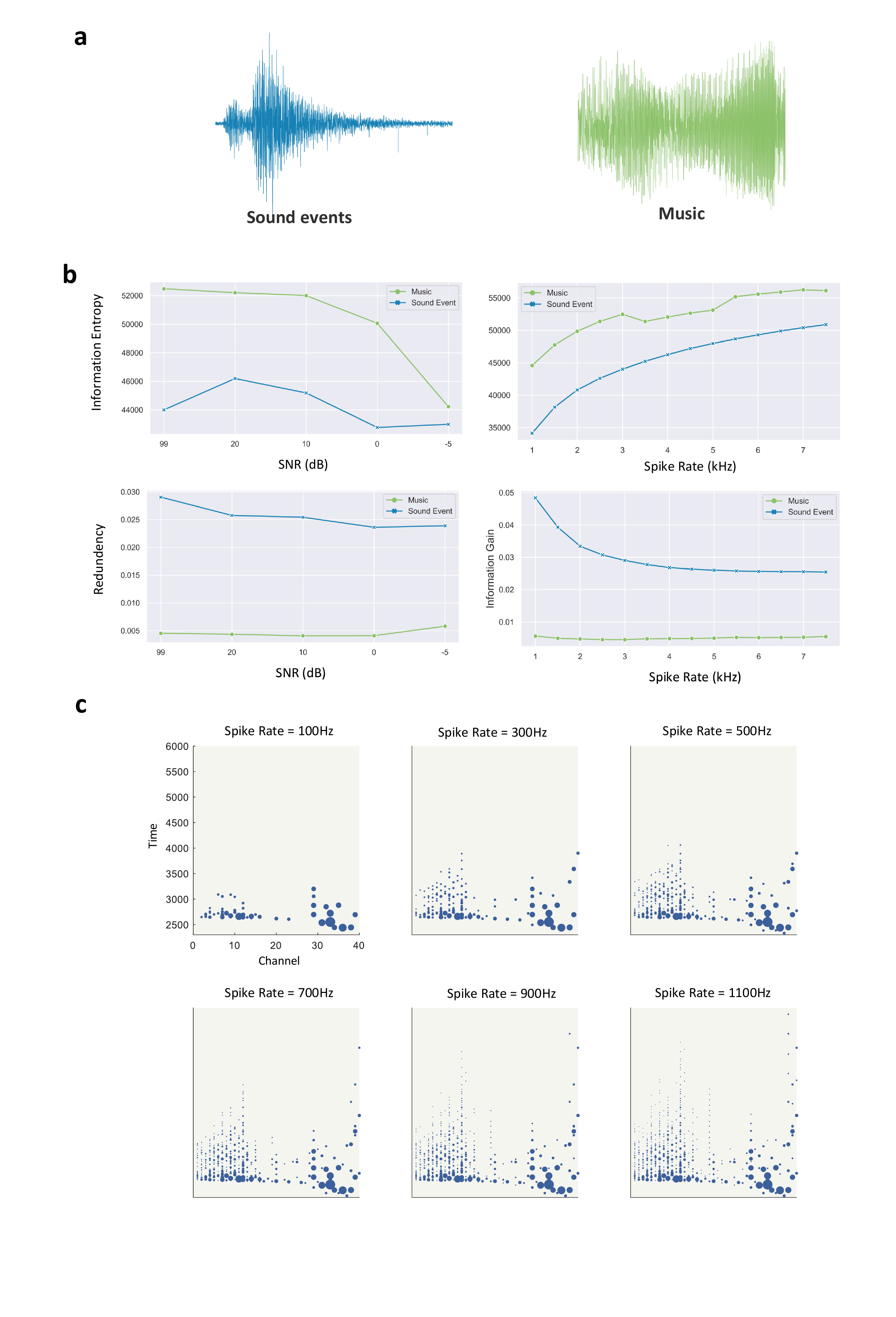}
	\caption{Measuring the amount of information in \emph{spiketrum} under different conditions. \textbf{a} Two categories of sounds (sound events and music) have different statistical regularities. \textbf{b} As the spike rate increases, the information entropy grows gradually, but with a decreasing rate (top). The relatively low redundancy indicates the capability of the \emph{spiketrums} to generate the sparse spike trains. (bottom). \textbf{c} Information entropies and redundancy are basically stable under different SNRs, $\lambda$=4kHz. \textbf{d} Structure invariance property of \emph{spiketrum} under different setting of spike rates. The size of the dots indicates the strength of intensities. When $\lambda$ is reduced from 1100Hz to 100 Hz, \emph{spiketrum} is able to retain the most principal structure in the spike pattern. }
	\label{fig.information}
\end{figure}

As the natural sound ensemble is a complex mixture of various sound categories, and different categories possess distinct statistical characteristics\cite{Smith2006Efficient}, we collect two common categories of sounds, including sound events, and music (Fig. \ref{fig.information}a; see Supplementary Materials) to explore the information entropies and redundancy of \emph{spiketrums}.
Let $H_c$ and $H_C$ denote the information entropy of independent neurons and neural population, respectively. As a function of $H_c$ and $H_C$, the redundancy $H_G$ defines the proportion of reduced uncertainty caused by the neural correlations, indicating the robustness to spike fluctuations.
\emph{Spiketrum} carries more information for higher spike rate $\lambda$ generally. Nevertheless, the growth trends of information entropy gradually slow down as $\lambda$ increases, implying that \emph{spiketrum} is inherently robust to the reduction of spikes (Fig. \ref{fig.information}b top).
Moreover, while the redundancy of different categories of sounds diverge when $\lambda$ is low (Fig. \ref{fig.information}b bottom, 0-3kHz),
they tend to be compressed closely when $\lambda$ is high enough.
Hence, the structure of \emph{spiketrum} has been stabilized for high $\lambda$, and the increase of $\lambda$ yields little effect on the sound representation, which is consistent with the representation precision analysis (Fig. \ref{fig.strategies_acc}).
Moreover, the stabilized redundancy of $H_G$ are relatively low (Fig. \ref{fig.information}b bottom), indicating the capability of the \emph{spiketrums} to generate the sparse spike trains.
The analysis results of robustness to environmental noises (see Supplementary Materials) show that the noises barely affect the information entropy of \emph{spiketrum}, as there is comparatively small change even at the level of 0dB of signal-to-noise ratio (SNR).
In addition, \emph{spiketrum} brings the considerable redundancy among different SNR levels (Fig. \ref{fig.information}b left-bottom), implying that the informational correlations among spike trains in \emph{spiketrum} are also robust to environmental noises.

The above analysis is in agreement with that \emph{spiketrum} is a population coding in which single spike trains only contain a small portion of auditory information.
Moreover, the auditory information is expressed by the joint activities of the neuron population\cite{Jazayeri2006Optimal}.
This self-correlated coding mechanism is advantageous in the sense that it can maintain the principle signal structure unaffected (Fig. \ref{fig.information}c) by destruction or noises. With the decrease of $\lambda$, the spiketrum encoding discards the spikes with minor projection (i.e., carrying less information) and retains the most important components in the original sound signal (see  E-TMP algorithm). As shown in Fig. \ref{fig.information}c, even when $\lambda$ is reduced to 100 spikes/second, spiketrum can still retain the
most crucial
structure in the spike pattern.
It is a highly desirable advantage for neuromorphic hardware\cite{davies2018loihi} which often faces limited on-board/on-chip resources, e.g., the spike transmission bandwidth, which would need the sensory encoding model to adapt $\lambda$ but not to distort the primary information.
\\
\begin{figure*}[htb] 
	\centering
\begin{center}
\includegraphics[width= .9\linewidth]{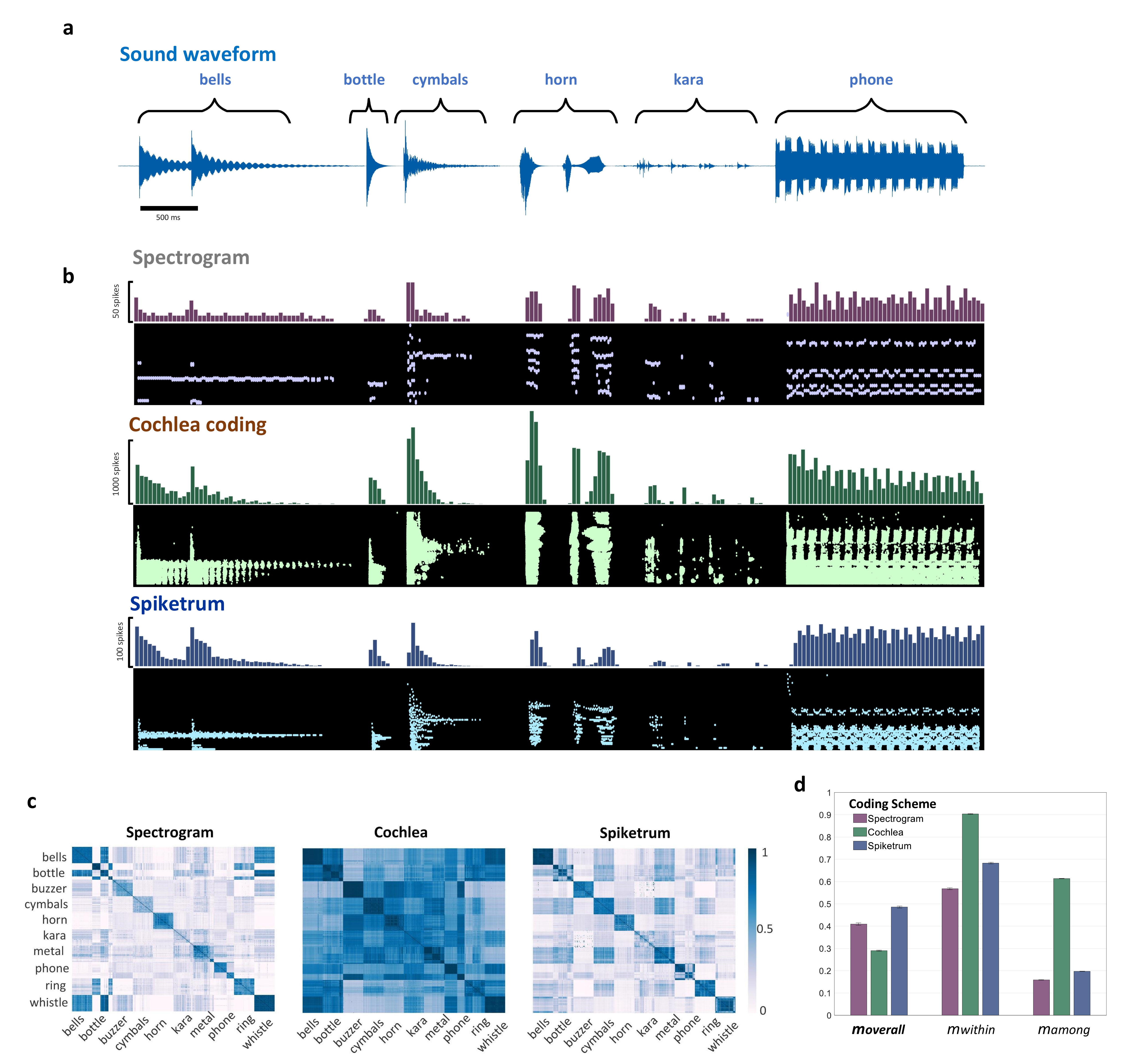}
\end{center}
\vspace{-10pt}
\caption{Discrimination performance of the three coding models. \textbf{a} The original sound waveform of different sound events. \textbf{b} The encoded spatiotemporal patterns by spiketrum. \textbf{c} The similarity matrices of three models. Each matrix illustrates the PSTH similarities between spike pattern pairs derived from 500 sound waveforms for each model. The 500 sound waveforms contain 10 categories of sound events and each category consists of 50 samples. \textbf{d} The average similarities of the within-class ($m_{within}$), among-class samples ($m_{among}$) and the overall scores ($m_{overall}$). The \emph{spiketrum} achieves higher discrimination performance ($m_{overall} = 0.4858$) than both competitors for CI (confidence level) = $95\%$.
}\label{fig.similarity}
\vspace{-5pt}
\end{figure*}

\subsection{ The Enhanced Discriminability of Auditory Patterns }  
Having shown that spiketrum encoding achieves sparse, precise and robust representation of auditory details, we are intrigued to verify whether its ability to represent sound details will enhance the discriminability of auditory inputs .

We collected a diversity of sound events\cite{Nakamura2000Acoustical} (bells, bottle, cymbals, horn, kara, phone) to construct a 4.7s long sound waveform (Fig. \ref{fig.similarity}a), and compute its peri-stimulus time histograms (PSTHs) and spike patterns (raster plots (Fig. \ref{fig.similarity}b).
The sound frequency perception range is set to 20Hz-8kHz. Further, we set $\lambda=1500Hz$ to achieve a trade-off between sparsity and representational effect according to Fig. \ref{fig.strategies_acc}b.
The spiketrum model is evaluated against the spectrogram-based model\cite{Dennis2015Combining} and the cochlea model\cite{Rudnicki2015Modeling}. As the two baseline models are not applicable to control the spike rates, we adopted their standard parameters, and their resulting average spike rates are 840Hz and 8kHz respectively.
As an image-based encoding method, the spectrogram model extracts the local energy peaks as features and encodes them into spike patterns, albeit roughly maintain the time-frequency structure of sound, it generates spike patterns too sparse and fails to describe the variation detail of spike rates.
Fig. \ref{fig.similarity}b shows that the time intervals between spikes are not precise enough to contain considerable details due to the block-based property of spectrogram in some areas (the purple plots for bells, bottle and cymbals).
The cochlea model integrates a selection of inner-ear models and produces spike patterns sharing high similarities with those from auditory neurons.
\begin{figure*}[htb!]
	\centering
	\includegraphics[width=\linewidth]{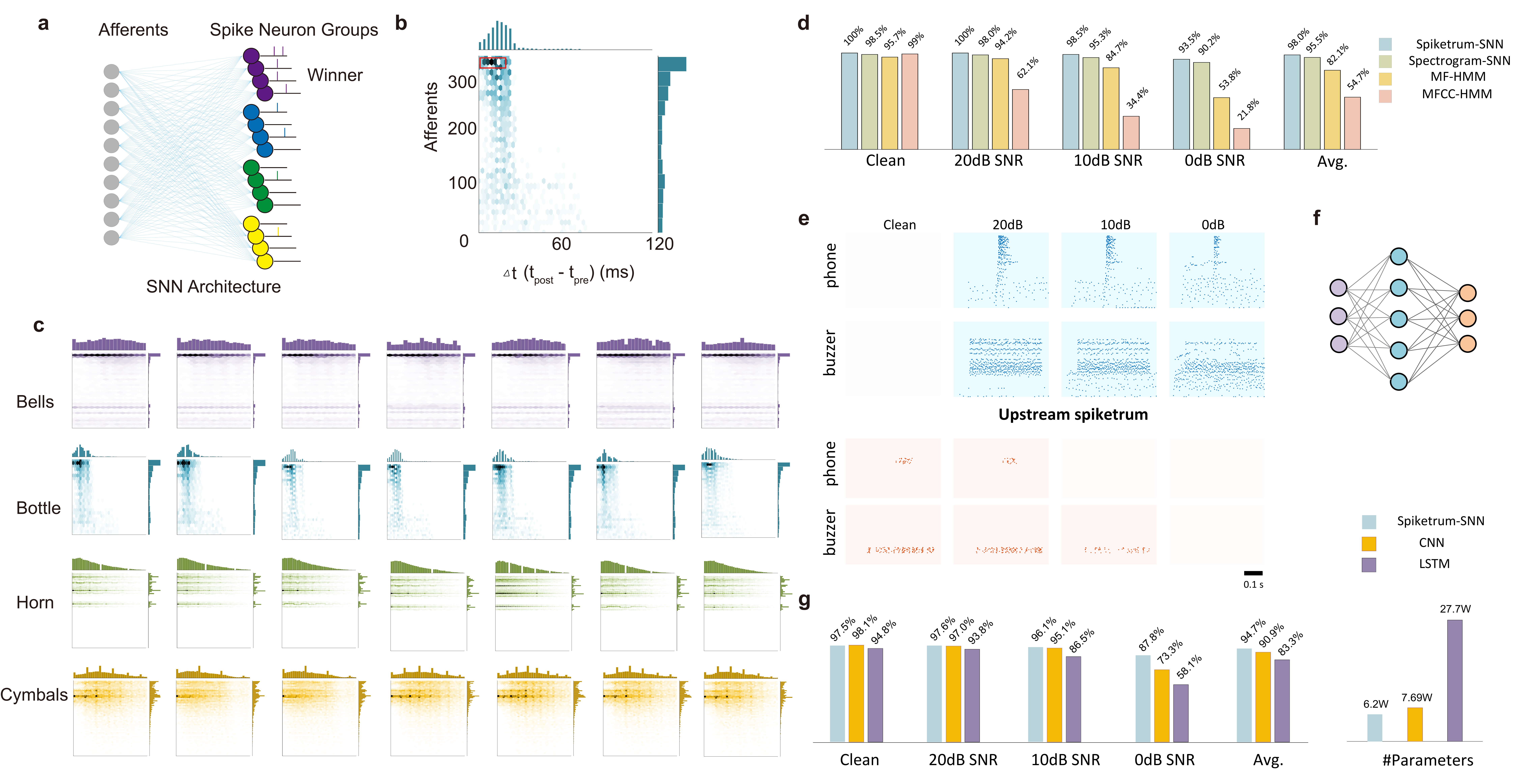}
	\caption{{Training an SNN (384-200) to distinguish 10 categories of sound events by learning their \emph{spiketrums}.
\textbf{a} The architecture of the single-layer SNN.
These neurons are grouped by their positive labels (10 groups corresponding to 10 classes of sound events; 20 neurons for each of the groups) and predict the label of the winner (the group firing the most spikes).  Specially, the synaptic outputs are called as "afferents" for downstream SNNs here
here.
} \textbf{b} The spike triggered average (STA) of a 'bottle' neuron illustrates the joint distribution (Afferents vs. $\Delta t$) of the pre-synaptic spikes which appears within the 120ms before the post-synaptic spike time. The darker the color in the pattern, the higher the probability that a pre-synaptic spike at the position will promote post neuron firing. So the spikes (red box area) issued by the corresponding afferent is most likely to stimulate this post neuron to fire within the next 10-20 ms. The histograms along top and right are the marginal distributions of the pre-synaptic spikes. \textbf{c} The STAs of 28 randomly selected neurons (4 classes, 7 neurons for each class).
\textbf{d} Classification results of SNN and machine learning models under different levels of background noise for 10 categories of sound events from RWCP dataset.
		\textbf{e} The downstream spiking neurons can learn the auditory features embedded in the upstream spiketrums. They transform the dense and background-corrupted spiketrums into sparse and background-invariant representations. Further, these neurons will decrease or stop firing in levels of background sound that preclude recognition. \textbf{f} The fully-connected architecture of the DSNN (128-450-10) for instruments recognition, using 10 different musical instrument classes from MedleyDB dataset.
		\textbf{g} By directly learning auditory features from spiketrums, SNN is able to achieve comparable performance with specialized ANNs, but with much fewer parameters and more straightforward structure.
	}
  \vspace{-4pt}
	\label{fig.sound_events}
\end{figure*}
However, simply mimicking the auditory peripheral coding mechanisms increases the computational complexity and makes the spike codes very dense and redundant. In contrast, \emph{spiketrum} shows a more transparent and refined spike distribution and thus can express temporal auditory structure precisely. Note that even though the spike rate of the cochlea model is much higher than \emph{spiketrum}, \emph{spiketrum} achieves a comparable representational effect on variation details (see the variation tendency of PSTH in Fig. \ref{fig.similarity}b).

We further characterize the discriminability via PSTH similarity measurement (see Supplementary Materials). Fig. \ref{fig.similarity}c presents the PSTH-similarity matrices of the three models. These matrices cover ten classes of sound events with fifty samples per class sampled from the RWCP dataset\cite{Nakamura2000Acoustical} (see Supplementary Materials). Analytically, the good discriminative ability for different categories of sounds results from the substantial gap between the within-class and among-class similarities. Thus, we compute the overall scores of the models $m_{overall}$, i.e., the difference between the average within-class similarity $m_{within}$ and the average among-class similarity $m_{among}$. The statistical results indicate that \emph{spiketrum} obtains the best discrimination performance, as the value of $m_{overall}$ is significantly higher than that of the other two models (Fig. \ref{fig.similarity}d).

\section{ {Spiketrum} Facilitates the Learning of SNNs}
We have shown that spiketrum can obtain sparse, robust and high precision spike codes for representing acoustic information, and we are intrigued to investigate whether these advantages in representation can indeed transform into benefits for training downstream spiking neural networks.
We utilize a simple supervised rule called \emph{tempotron}\cite{G2006The} to train a single-layer SNN constructed by leaky integrate-and-fire (LIF) neurons (Fig. \ref{fig.sound_events}a).
We analyze the spike-triggered average (STA) of the neurons trained to respond to the sound events and also joint distributions with respect to $\Delta t$ (the difference between the post-synaptic spike time $t_{post}$ and the pre-synaptic spike time $t_{pre}$) and the positions of pre-synaptic afferents (Fig. \ref{fig.sound_events}b).
The darker the color of a point, the higher the probability that a pre-synaptic spike at the position will promote post neuron to fire.
Therefore, the learned auditory neural feature by \emph{Spiketrum} could facilitate the trained neuron itself to fire.
Further, we randomly select 28 trained neurons (4 classes, 7 neurons of each class) and visualize their STAs (Fig. \ref{fig.sound_events}c).
The patterns corresponding to the same class show roughly similar characteristics, and some classes develop multiple features.
By contrast, the patterns of different classes appear distinctly different in spike distributions.
The STA results demonstrate that that neurons can learn discriminative auditory features from \emph{spiketrum}, and spatiotemporal distributions of spikes can reflect the time-frequency modulation of neurons.
To explore the robustness of different models against environmental noises,
we add different levels of babble noise onto the clean testing individual sound events and evaluate the performance of these models on these background-corrupted sounds. The recognition accuracy of SNNs using spiketrum is higher than traditional auditory models (MF-HMM and MFCC-HMM) in all conditions (Fig. \ref{fig.sound_events}d), showing that learning of SNNs is minimally affected by background noises.
Noticeably, spiketrum as inputs enable SNNs to transform the dense representations from encoding neurons into the sparse and selective firing patterns of downstream learning neurons, which will decrease or stop firing as SNR decreases (Fig. \ref{fig.sound_events}e). This
phenomenon
is consistent with the behavior of auditory neurons found in the zebra finch\cite{Schneider2013Sparse}.

To further demonstrate that \emph{spiketrum} is able to facilitate spiking neurons to learn more complex auditory features, we compare a simple deep spiking neural network (DSNN) (Fig. \ref{fig.sound_events}f; see Supplementary Materials) with non-spiking deep neural networks (DNNs) by performing music timbre recognition tasks. In addition to pitch, loudness, and duration, timbre is a set of complex acoustic attributes associated with spectral and temporal structures embedded in sounds\cite{Pons2017Timbre}.

Fig. \ref{fig.sound_events}g shows that SNNs using spiketrum achieve competitive performance to typical deep neural networks with fewer parameters and simpler structure on that task without any noise, which is highly beneficial to hardware implementation.
Moreover, the spiketrum with SNNs is more robust than the direct spectrogram method with CNNs and LSTMs models under the different noise levels from 0dB to 20dB. There is only the accuracy drop of $9.7\%$ under 0 db noise environment for spiketrum with SNNs, which is much lower than the ones of $24.8 \%$ and $36.7 \%$ for CNNs and LSTM, respectively.
It seems comprehensible as the idea which takes the spectrogram as current inputs from different spectrum channels directly doesn't encode sounds using spikes with necessary information loss, while our spiketrum preserve enough information by minimizing the information loss during the waveform-to-spike transformation. Hence our spiketrum is more robust to neural fluctuation and spike losses under different noise levels.

\section{ Neuromorphic Cochlea Hardware}

The outstanding ability of the proposed algorithm to encode information efficiently using a small number of spike activities (providing low communication and processing bandwidth) with high accuracy (can reconstruct the original signals accurately by decoding output spikes) is further verified at the hardware level while investigating how well the algorithm translates into an efficient real-time hardware implementation.

\begin{figure}[t] 
   \centering
   \includegraphics[width= \linewidth]{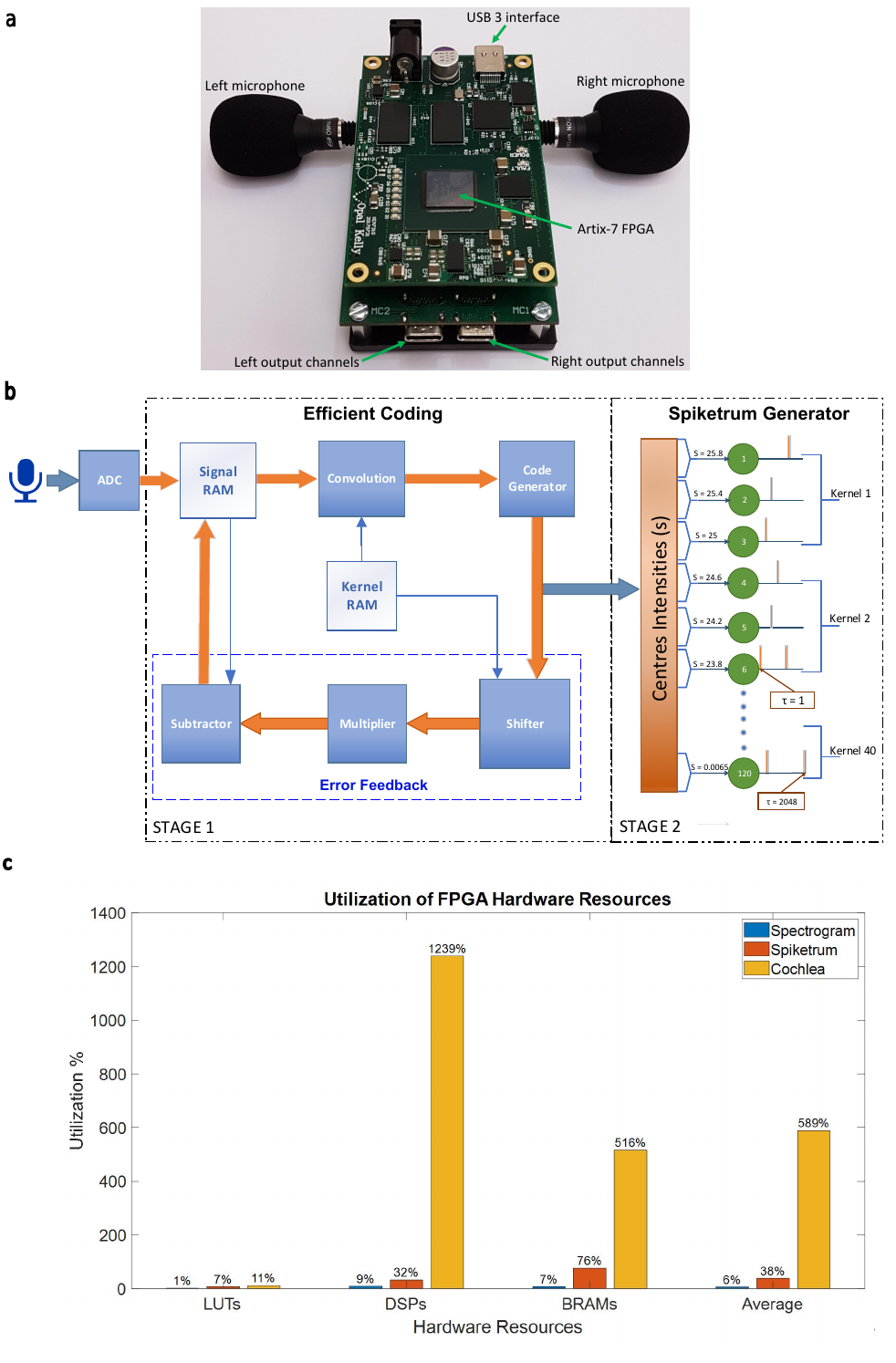}
   \caption{
         The neuromorphic spiketrum cochleae implementation based on FPGA. \textbf{a} Prototype of real-time neuromorphic cochlea and architecture design. An audio input stream can be provided from two microphones or via a USB 3 interface (using an external device). XEM7310 Xilinx Artix-7 FPGA board process incoming audio signals in real-time. The 120 spike output channels from each cochlea are available via a USB-C interface.
         \textbf{b} Architecture design of the spiketrum cochlea. The input signal received from the microphone is converted
to a digital signal using the ADC. At stage 1, the Error Feedback block facilitates generation of subsequent codes iteratively to minimize
the error of the encoded representation of the incoming audio signal. Each time a code is generated at the output of
the Code Generator, the ITP coding scheme at stage 2 generates a spike in the output channel of the cochlea.}
   \label{fig.cochlea_diagram}
 \end{figure}

The hardware hearing device has two cochleae implementations. Each cochlea has 120 spike output channels corresponding to three discrete intensity levels (called characteristic intensities) for each of the 40 Gammatone kernels. The hardware generates codes following the steps in the algorithm, and each time when a code ($m_i$,$\tau_i$,$s_i$) is generated, the ITP coder is activated to produce a spike at the temporal location $\tau_i$ on the channel corresponding to the kernel index $m_i$ and the characteristic intensity level which is nearest to the intensity value of the code $s_i$ (see the block diagram of the FPGA in Fig. \ref{fig.cochlea_diagram}b).

The neuromorphic cochlea prototype is implemented in an XEM7310 Xilinx Artix-7 FPGA board which integrates
SDRAMs, EEPROMs and clock generators. As shown in Fig. \ref{fig.cochlea_diagram}b, the
\textbf{red} {analog}
 signal from the microphone is
converted to a digital signal with a sampling rate of 16 kHz, using an Analogue-to-Digital converter (ADC)
and stored on the Signal Random Access Memory (RAM). Then the time-domain convolution is performed between
the digital audio signal and the 40 kernels by multiplying the Fast Fourier transformed signal and the kernels
in the frequency-domain and obtaining the Inverse Fast Fourier transform of results to speed up calculations.
The frequency-domain kernel data is stored in the Kernel RAM to
minimize
 real-time computations. To facilitate real-time operation, at a given time, the hardware processes a buffered segment of the continuously sampled incoming audio signal. To provide the optimum balance between the use of available hardware resources on the FPGA and encode efficiency, the length of the segment is chosen as 43.5 $ms$. The output
of the convolution indicates a degree of correlation of the audio segment with each
centered-frequency kernel.
Using the coordinates of the maximum value of the convoluted output (i.e., the value corresponding to the most correlated kernel, the $i_{th}$ kernel), the Code Generator block generates the code, which produces the kernel index $m_i$, temporal position $\tau_i$ and intensity of the $i_{th}$ kernel $s_i$.

Fig. \ref{fig.cochlea_diagram}b illustrates the hardware implementation of the Error Feedback using a block diagram. The Error Feedback first calculates the portion of the signal represented by the recently generated code, $s_i \phi_{m_i}(t-\tau_i)$, by shifting $\phi_{m_i}(t)$ to $\phi_{m_i}(t-\tau_i)$ and multiplying it with $s_i$. Then, it is subtracted from the signal $x(t)$ to obtain the non-represented portion (residual) of the signal, $x(t)_{new}$ (i.e., $x(t)_{new}=x(t)- s_i \phi_{m_i}(t-\tau_i)$), and finally, the Signal RAM is updated with the new signal,  $x(t)_{new}$. Once the Signal RAM is updated using the Error Feedback block, Convolution and Code Generation blocks generate subsequent code. This process continues iteratively to generate more codes to represent the incoming audio signal accurately. However, not to violate the real-time operation of the cochlea, the process could only iterate until a new segment of the input signal is available to be processed.


\begin{figure}[t]
		\centering
		\includegraphics[width=\linewidth]{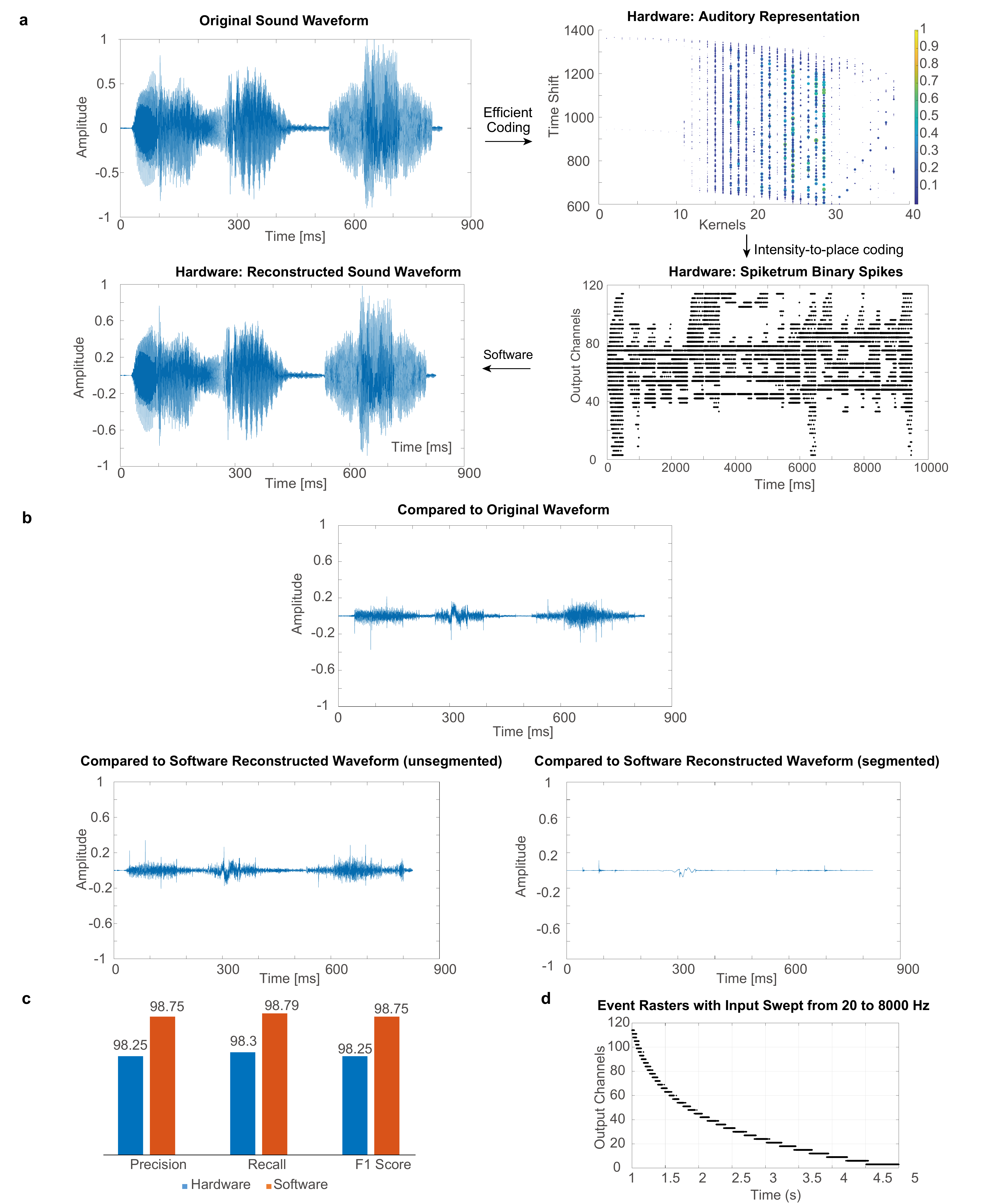}
		\caption{
        	The representation and error evaluation of the hardware implementation of Spiketrum.  \textbf{a} Shows the hardware encoded spiketrum auditory representation of an input sound waveform, generation of spike trains using hardware intensity-to-place coding and reconstruction of the hardware-encoded spike train using a software decoder to demonstrate the similarity of the constructed waveform with the original sound waveform. \textbf{b} Error graphs for the hardware output. Top: the difference between the original sound waveform and the hardware reconstructed waveform; Bottom Left: the difference between the software and hardware reconstructed waveforms; Bottom Right: the difference between the software and hardware reconstructed waveforms based on the segmented method. \textbf{c} shows the ability to perform SNN learning and classifications tasks compared to the software implementation.
        	The SNN classification is done by directly learning auditory features from hardware and software implementations independently to produce comparable performance;
        	Here, five different musical instrument classes from MedleyDB\cite{Bittner2014MedleyDB} dataset, namely, acoustic guitar, cello, clarinet, cymbal and double bass, were encoded; Each class contains 160 sound waveforms with different melodies and styles. These waveforms are 1.5s long and sampled at 16 kHz.
        	\textbf{d} shows the output response of the cochlea, using the raster plot of the 120 output channels, to an input of audio signal frequency linearly varying from 20 Hz to 8 kHz within 5 seconds.}
		\label{fig.cochlea_performance}
\end{figure}

The comparison of the spiketrum hardware implementation with the spectrogram-based and cochlea-based implementations shows that the spiketrum has a good balance between the biological plausibility and the ability to translate into an efficient real-time hardware implementation. Based on the estimated values, the cochlea-based method requires a considerable amount of hardware resources and power to achieve similar precision and latency as those of spiketrum implementation, costing about 15 times more hardware resources and 25 times more average power consumption (Fig. \ref{fig.cochlea_diagram}c). The spiketrum model generates sparse spike trains which reduce the activity-based computations needed at the subsequent processing stages. In particular, the output of the spiketrum can easily decode to reconstruct the original format of the signal using a lossless reconstruction algorithm, providing an accurate evaluation of the encoding performance of the hardware cochlea, this is not the case with phenomenological models such as spectrogram-based.

The prototype operates in real-time with a 200 MHz clock speed to produce up to 2,000 spikes per second per cochlea to encode incoming acoustic signals sampled at 16 kHz. Hardware processing latency is 0.5 ms. The characteristics curve of the hardware cochlea is provided in Fig. \ref{fig.cochlea_performance}d, demonstrating the filtering ability of the cochlea across the 20 Hz to 8 kHz frequency spectrum of sound. As shown in Fig. \ref{fig.cochlea_performance}a, the hearing device accurately generates the spiketrum, closely following the computations performed in the software algorithm. The ability to learn and classify the spike outputs generated by the hardware and software spiketrums with high accuracy is demonstrated in Fig. \ref{fig.cochlea_performance}c, using a typical example, where the classifications are performed by feeding five categories of sound events to an SNN via spiketrums.
\section{Conclusion}

We have developed an efficient auditory neural coding model called \emph{spiketrum} that achieves compelling performance such as sparse, precise and robust representation of sensory signals, and provides a viable sensory encoding methodology for implementing spiking neural networks. We also presented a neuromorphic cochlea prototype that implemented the spiketrum model in real time.
Spiketrum provides a good balance between the level of abstraction of the biological cochlea and the ability to generate neural spikes, while reducing the computational load in hardware.
The proposed implementation can be translated into an Integrated Circuit (IC) using the latest ultra-low-power digital VLSI technology, where the enormous power saving and device size advantages can be achieved. 
The proposed sensory technology can help perform complex day-to-day interactive intelligent tasks in spike-based intelligent computing machines.

   \renewcommand{\figurename}{Supplementary Fig.}
  \begin{figure*}[H]
    \centering
    \includegraphics[width= \linewidth]{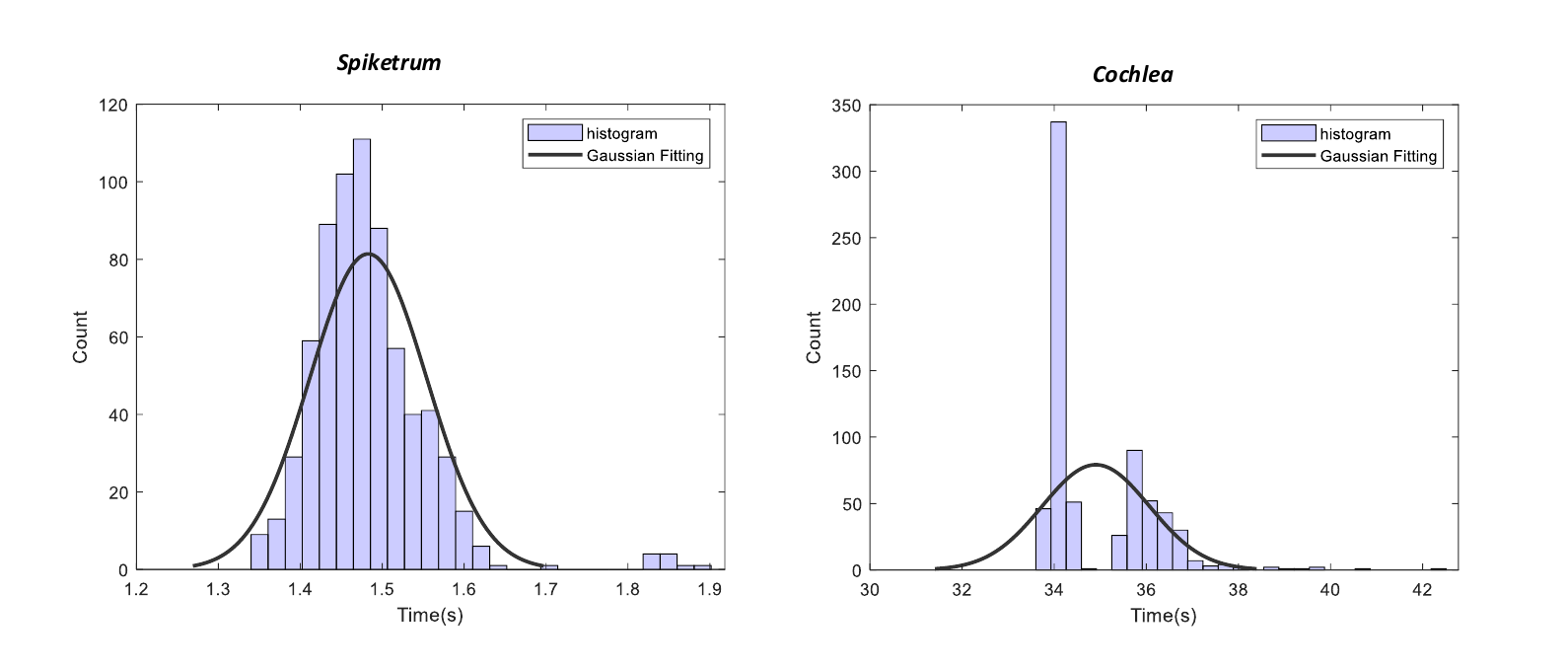}
    \caption{
      The
      running time comparison of the spiketrum and the cochlea-based method on the CPUs.
    We test the running time of the E-TMP module on 700 pieces of 1.5s sound clips randomly sampled from MedlyDB with a
     spike rate of 1000 Hz.
      Noticeably, the early stopping mechanism is adopted when the representation precision $P$ is larger than $99\%$. The result shows spiketrum obtains $23\times$
    gain on running time than the Cochlea-based method\cite{cramer2020heidelberg} approximately. Furthermore, the overall range of time
    overhead of spiketrum is $[1.2689,  1.6970]$ based on the three-
    sigma rule. So, it just needs nearly the same time to process a 1.5s clip sampled
  at 16 kHz with PCs which could be further improved
     from different dimensions such as segment-based optimization, concurrence programming, neuromorphic hardware acceleration,
  etc.
    }

    \label{fig.time_cost}
   \end{figure*}
\bibliographystyle{IEEEtran}
\bibliography{main}

\begin{thebibliography}{10}
\providecommand{\url}[1]{#1}
\csname url@samestyle\endcsname
\providecommand{\newblock}{\relax}
\providecommand{\bibinfo}[2]{#2}
\providecommand{\BIBentrySTDinterwordspacing}{\spaceskip=0pt\relax}
\providecommand{\BIBentryALTinterwordstretchfactor}{4}
\providecommand{\BIBentryALTinterwordspacing}{\spaceskip=\fontdimen2\font plus
\BIBentryALTinterwordstretchfactor\fontdimen3\font minus
  \fontdimen4\font\relax}
\providecommand{\BIBforeignlanguage}[2]{{%
\expandafter\ifx\csname l@#1\endcsname\relax
\typeout{** WARNING: IEEEtran.bst: No hyphenation pattern has been}%
\typeout{** loaded for the language `#1'. Using the pattern for}%
\typeout{** the default language instead.}%
\else
\language=\csname l@#1\endcsname
\fi
#2}}
\providecommand{\BIBdecl}{\relax}
\BIBdecl

\bibitem{pei2019towards}
J.~Pei, L.~Deng, S.~Song, M.~Zhao, Y.~Zhang, S.~Wu, G.~Wang, Z.~Zou, Z.~Wu,
  W.~He \emph{et~al.}, ``Towards artificial general intelligence with hybrid
  tianjic chip architecture,'' \emph{Nature}, vol. 572, no. 7767, pp. 106--111,
  2019.

\bibitem{merolla2014million}
P.~A. Merolla, J.~V. Arthur, R.~Alvarez-Icaza, A.~S. Cassidy, J.~Sawada,
  F.~Akopyan, B.~L. Jackson, N.~Imam, C.~Guo, Y.~Nakamura \emph{et~al.}, ``A
  million spiking-neuron integrated circuit with a scalable communication
  network and interface,'' \emph{Science}, vol. 345, no. 6197, pp. 668--673,
  2014.

\bibitem{furber2014spinnaker}
S.~B. Furber, F.~Galluppi, S.~Temple, and L.~A. Plana, ``The spinnaker
  project,'' \emph{Proceedings of the IEEE}, vol. 102, no.~5, pp. 652--665,
  2014.

\bibitem{schemmel2010wafer}
J.~Schemmel, D.~Briiderle, A.~Griibl, M.~Hock, K.~Meier, and S.~Millner, ``A
  wafer-scale neuromorphic hardware system for large-scale neural modeling,''
  in \emph{IEEE International Symposium on Circuits and Systems}, 2010, pp.
  1947--1950.

\bibitem{Giacomo2015}
G.~Indiveri and S.-C. Liu, ``Memory and information processing in neuromorphic
  systems,'' \emph{Proceedings of the IEEE}, vol. 103, no.~8, pp. 1379--1397,
  2015.

\bibitem{Abbott2000Synaptic}
L.~F. Abbott and S.~B. Nelson, ``Synaptic plasticity: taming the beast.''
  \emph{Nature Neuroscience}, vol.~3, no. 11s, p. 1178, 2000.

\bibitem{song2000competitive}
S.~Song, K.~D. Miller, and L.~F. Abbott, ``Competitive hebbian learning through
  spike-timing-dependent synaptic plasticity,'' \emph{Nature Neuroscience},
  vol.~3, no.~9, p. 919, 2000.

\bibitem{Maass2015}
W.~Maass, ``To spike or not to spike: That is the question,'' \emph{Proceedings
  of the IEEE}, vol. 103, no.~12, pp. 2219--2224, 2015.

\bibitem{Mem2014}
R.-M. Memmesheimer, R.~Rubin, B.~P. {\"O}lveczky, and H.~Sompolinsky,
  ``Learning precisely timed spikes,'' \emph{Neuron}, vol.~82, no.~4, pp.
  925--938, 2014.

\bibitem{Lillicrap2016Random}
T.~P. Lillicrap, D.~Cownden, D.~B. Tweed, and C.~J. Akerman, ``Random synaptic
  feedback weights support error backpropagation for deep learning,''
  \emph{Nature Communications}, vol.~7, p. 13276, 2016.

\bibitem{G2016Spiking}
R.~Gütig, ``Spiking neurons can discover predictive features by
  aggregate-label learning,'' \emph{Science}, vol. 351, no. 6277, p. aab4113,
  2016.

\bibitem{Nicola2017Supervised}
W.~Nicola and C.~Clopath, ``Supervised learning in spiking neural networks with
  force training,'' \emph{Nature Communications}, vol.~8, no.~1, p. 2208, 2017.

\bibitem{benjamin2014neurogrid}
B.~V. Benjamin, P.~Gao, E.~McQuinn, S.~Choudhary, A.~R. Chandrasekaran, J.-M.
  Bussat, R.~Alvarez-Icaza, J.~V. Arthur, P.~A. Merolla, and K.~Boahen,
  ``Neurogrid: A mixed-analog-digital multichip system for large-scale neural
  simulations,'' \emph{Proceedings of the IEEE}, vol. 102, no.~5, pp. 699--716,
  2014.

\bibitem{davies2018loihi}
M.~Davies, N.~Srinivasa, T.-H. Lin, G.~Chinya, Y.~Cao, S.~H. Choday, G.~Dimou,
  P.~Joshi, N.~Imam, S.~Jain \emph{et~al.}, ``Loihi: A neuromorphic manycore
  processor with on-chip learning,'' \emph{IEEE Micro}, vol.~38, no.~1, pp.
  82--99, 2018.

\bibitem{Liu2016}
M.~Yang, C.-H. Chien, T.~Delbruck, and S.-C. Liu, ``A 0.5 v 55 $\mu \text{W}$
  64 $\times $ 2 channel binaural silicon cochlea for event-driven stereo-audio
  sensing,'' \emph{IEEE Journal of Solid-State Circuits}, vol.~51, no.~11, pp.
  2554--2569, 2016.

\bibitem{Qiao2015}
N.~Qiao, H.~Mostafa, F.~Corradi, M.~Osswald, F.~Stefanini, D.~Sumislawska, and
  G.~Indiveri, ``A reconfigurable on-line learning spiking neuromorphic
  processor comprising 256 neurons and 128k synapses,'' \emph{Frontiers in
  Neuroscience}, vol.~9, p. 141, 2015.

\bibitem{Tenore2011Neuromorphic}
F.~Tenore and R.~Etienne-Cummings, ``Neuromorphic electronic design,''
  \emph{Biohybrid Systems: Nerves, Interfaces, and Machines}, pp. 31--51, 2011.

\bibitem{cao2015spiking}
Y.~Cao, Y.~Chen, and D.~Khosla, ``Spiking deep convolutional neural networks
  for energy-efficient object recognition,'' \emph{International Journal of
  Computer Vision}, vol. 113, no.~1, pp. 54--66, 2015.

\bibitem{Shrestha2018}
S.~B. Shrestha and G.~Orchard, ``{SLAYER}: Spike layer error reassignment in
  time,'' in \emph{Advances in Neural Information Processing Systems 31},
  S.~Bengio, H.~Wallach, H.~Larochelle, K.~Grauman, N.~Cesa-Bianchi, and
  R.~Garnett, Eds.\hskip 1em plus 0.5em minus 0.4em\relax Curran Associates,
  Inc., 2018, pp. 1419--1428.

\bibitem{Emre2019}
E.~O. Neftci, H.~Mostaf, and F.~Zenke, ``Surrogate gradient learning in spiking
  neural networks: Bringing the power of gradient-based optimization to spiking
  neural networks,'' \emph{IEEE Signal Processing Magazine}, vol.~36, no.~6,
  pp. 51--63, 2019.

\bibitem{roy2019towards}
K.~Roy, A.~Jaiswal, and P.~Panda, ``Towards spike-based machine intelligence
  with neuromorphic computing,'' \emph{Nature}, vol. 575, no. 7784, pp.
  607--617, 2019.

\bibitem{Ding2012Neural}
N.~Ding and J.~Z. Simon, ``Neural coding of continuous speech in auditory
  cortex during monaural and dichotic listening,'' \emph{Journal of
  Neurophysiology}, vol. 107, no.~1, pp. 78--89, 2012.

\bibitem{Schneider2013Sparse}
D.~M. Schneider and S.~M.~N. Woolley, ``Sparse and background-invariant coding
  of vocalizations in auditory scenes.'' \emph{Neuron}, vol.~79, no.~1, pp.
  141--152, 2013.

\bibitem{Oxenham2018How}
A.~J. Oxenham, ``How we hear: The perception and neural coding of sound,''
  \emph{Annual Review of Psychology}, vol.~69, no.~1, p.~27, 2018.

\bibitem{Rhode1994Encoding}
W.~S. Rhode and S.~Greenberg, ``Encoding of amplitude modulation in the
  cochlear nucleus of the cat.'' \emph{Journal of Neurophysiology}, vol.~71,
  no.~5, pp. 1797--1825, 1994.

\bibitem{Krishna2000Auditory}
B.~S. Krishna and M.~N. Semple, ``Auditory temporal processing: responses to
  sinusoidally amplitude-modulated tones in the inferior colliculus.''
  \emph{Journal of Neurophysiology}, vol.~84, no.~1, p. 255, 2000.

\bibitem{Bartlett2007Neural}
E.~L. Bartlett and X.~Wang, ``Neural representations of temporally modulated
  signals in the auditory thalamus of awake primates.'' \emph{Journal of
  Neurophysiology}, vol.~97, no.~2, pp. 1005--1017, 2007.

\bibitem{Sharpee2011Hierarchical}
T.~O. Sharpee, C.~A. Atencio, and C.~E. Schreiner, ``Hierarchical
  representations in the auditory cortex,'' \emph{Current Opinion in
  Neurobiology}, vol.~21, no.~5, pp. 761--767, 2011.

\bibitem{Dennis2015Combining}
J.~Dennis, H.~D. Tran, and H.~Li, ``Combining robust spike coding with spiking
  neural networks for sound event classification,'' in \emph{IEEE International
  Conference on Acoustics, Speech and Signal Processing}, 2015, pp. 176--180.

\bibitem{Sompolinsky2009Time}
H.~Sompolinsky, ``Time-warp-invariant neuronal processing.'' \emph{Plos
  Biology}, vol.~7, no.~7, p. e1000141, 2009.

\bibitem{Schafer2014Noise}
P.~B. Schafer and D.~Z. Jin, ``Noise-robust speech recognition through auditory
  feature detection and spike sequence decoding,'' \emph{Neural Computation},
  vol.~26, no.~3, pp. 523--556, 2014.

\bibitem{singh2003modulation}
N.~C. Singh and F.~E. Theunissen, ``Modulation spectra of natural sounds and
  ethological theories of auditory processing,'' \emph{The Journal of the
  Acoustical Society of America}, vol. 114, no.~6, pp. 3394--3411, 2003.

\bibitem{Tavanaei2016A}
A.~Tavanaei and A.~S. Maida, ``A spiking network that learns to extract spike
  signatures from speech signals,'' \emph{Neurocomputing}, vol. 240, pp.
  191--199, 2016.

\bibitem{Wang2008Neural}
X.~Wang, T.~Lu, D.~Bendor, and E.~Bartlett, ``Neural coding of temporal
  information in auditory thalamus and cortex,'' \emph{Neuroscience}, vol. 154,
  no.~1, pp. 294--303, 2008.

\bibitem{Elhilali2004Dynamics}
M.~Elhilali, J.~B. Fritz, D.~J. Klein, J.~Z. Simon, and S.~A. Shamma,
  ``Dynamics of precise spike timing in primary auditory cortex.''
  \emph{Journal of Neuroscience the Official Journal of the Society for
  Neuroscience}, vol.~24, no.~5, pp. 1159--72, 2004.

\bibitem{Schnupp2006Plasticity}
J.~W.~H. Schnupp, T.~M. Hall, R.~F. Kokelaar, and B.~Ahmed, ``Plasticity of
  temporal pattern codes for vocalization stimuli in primary auditory cortex.''
  \emph{Journal of Neuroscience}, vol.~26, no.~18, pp. 4785--4795, 2006.

\bibitem{Araki2016Mind}
M.~Araki, M.~M. Bandi, and Y.~Yazaki-Sugiyama, ``Mind the gap: Neural coding of
  species identity in birdsong prosody,'' \emph{Science}, vol. 354, no. 6317,
  pp. 1282--1287, 2016.

\bibitem{Rosen1992Temporal}
S.~Rosen, ``Temporal information in speech: acoustic, auditory and linguistic
  aspects.'' \emph{Philos Trans R Soc Lond B Biol Sci}, vol. 336, no. 1278, pp.
  367--373, 1992.

\bibitem{holmberg2007speech}
M.~Holmberg, ``Speech encoding in the human auditory periphery: Modeling and
  quantitative assessment by means of automatic speech recognition,'' \emph{VDI
  Verlag}, 2009.

\bibitem{Darwin2005Pitch}
C.~J. Darwin, ``"pitch and auditory grouping",'' in \emph{Pitch: Neural Coding
  and Perception}.\hskip 1em plus 0.5em minus 0.4em\relax Springer New York,
  2005, pp. 278--305.

\bibitem{Furukawa2002Cortical}
S.~Furukawa and J.~C. Middlebrooks, ``Cortical representation of auditory
  space: information-bearing features of spike patterns,'' \emph{Journal of
  Neurophysiology}, vol.~87, no.~4, pp. 1749--1762, 2002.

\bibitem{zilany2014updated}
M.~S. Zilany, I.~C. Bruce, and L.~H. Carney, ``Updated parameters and expanded
  simulation options for a model of the auditory periphery,'' \emph{The Journal
  of the Acoustical Society of America}, vol. 135, no.~1, pp. 283--286, 2014.

\bibitem{Rudnicki2015Modeling}
M.~Rudnicki, O.~Schoppe, M.~Isik, F.~Völk, and W.~Hemmert, ``Modeling auditory
  coding: from sound to spikes,'' \emph{Cell \& Tissue Research}, vol. 361,
  no.~1, pp. 159--175, 2015.

\bibitem{leong2003fpga}
M.-P. Leong, C.~T. Jin, and P.~H. Leong, ``An fpga-based electronic cochlea,''
  \emph{EURASIP Journal on Advances in Signal Processing}, vol. 2003, no.~7, p.
  751437, 2003.

\bibitem{Liu2010Neuromorphic}
S.~C. Liu and T.~Delbruck, ``Neuromorphic sensory systems.'' \emph{Current
  Opinion in Neurobiology}, vol.~20, no.~3, pp. 288--295, 2010.

\bibitem{liu2014event}
S.-C. Liu, T.~Delbruck, G.~Indiveri, A.~Whatley, and R.~Douglas, ``Event-based
  neuromorphic systems,'' \emph{John Wiley \& Sons}, 2014.

\bibitem{Smith2006Efficient}
E.~C. Smith and M.~S. Lewicki, ``Efficient auditory coding.'' \emph{Nature},
  vol. 439, no. 7079, pp. 978--82, 2006.

\bibitem{Davis1997Adaptive}
G.~Davis, S.~Mallat, and M.~Avellaneda, ``Adaptive greedy approximations,''
  \emph{Constructive Approximation}, vol.~13, no.~1, pp. 57--98, 1997.

\bibitem{Krstulovic2011Mptk}
S.~Krstulovic and R.~Gribonval, ``Mptk: Matching pursuit made tractable,'' in
  \emph{IEEE International Conference on Acoustics Speech and Signal Processing
  Proceedings}, vol.~3.\hskip 1em plus 0.5em minus 0.4em\relax IEEE, 2006, pp.
  III--III.

\bibitem{Time2004}
M.~Abeles, ``Time is precious,'' \emph{Science}, vol. 304, no. 5670, pp.
  523--524, 2004.

\bibitem{flyScience}
S.~Dasgupta, C.~F. Stevens, and S.~Navlakha, ``A neural algorithm for a
  fundamental computing problem,'' \emph{Science}, vol. 358, no. 6364, pp.
  793--796, 2017.

\bibitem{oster2004winner}
M.~Oster and S.-C. Liu, ``A winner-take-all spiking network with spiking
  inputs,'' in \emph{Proceedings of the 2004 11th IEEE International Conference
  on Electronics, Circuits and Systems, 2004. ICECS 2004.}\hskip 1em plus 0.5em
  minus 0.4em\relax IEEE, 2004, pp. 203--206.

\bibitem{dean2005neural}
I.~Dean, N.~S. Harper, and D.~McAlpine, ``Neural population coding of sound
  level adapts to stimulus statistics,'' \emph{Nature neuroscience}, vol.~8,
  no.~12, pp. 1684--1689, 2005.

\bibitem{Lewicki2002Efficient}
M.~S. Lewicki, ``Efficient coding of natural sounds,'' \emph{Nature
  Neuroscience}, vol.~5, no.~4, pp. 356--363, 2002.

\bibitem{Valero2012Gammatone}
X.~Valero and F.~Alias, ``Gammatone cepstral coefficients: Biologically
  inspired features for non-speech audio classification,'' \emph{IEEE
  Transactions on Multimedia}, vol.~14, no.~6, pp. 1684--1689, 2012.

\bibitem{Normanhaignere2015Distinct}
S.~Normanhaignere, N.~G. Kanwisher, and J.~H. Mcdermott, ``Distinct cortical
  pathways for music and speech revealed by hypothesis-free voxel
  decomposition.'' \emph{Neuron}, vol.~88, no.~6, pp. 1281--1296, 2015.

\bibitem{Jazayeri2006Optimal}
M.~Jazayeri and J.~A. Movshon, ``Optimal representation of sensory information
  by neural populations,'' \emph{Nature Neuroscience}, vol.~9, no.~5, p. 690,
  2006.

\bibitem{Urbanczik2009Reinforcement}
R.~Urbanczik and W.~Senn, ``Reinforcement learning in populations of spiking
  neurons.'' \emph{Nature Neuroscience}, vol.~12, no.~3, pp. 250--252, 2009.

\bibitem{mesgarani2014mechanisms}
N.~Mesgarani, S.~V. David, J.~B. Fritz, and S.~A. Shamma, ``Mechanisms of noise
  robust representation of speech in primary auditory cortex,''
  \emph{Proceedings of the National Academy of Sciences USA}, vol. 111, no.~18,
  pp. 6792--6797, 2014.

\bibitem{Shannon1948A}
C.~E. Shannon, ``A mathematical theory of communication,'' \emph{Bell system
  technical journal}, vol.~27, no.~3, pp. 379--423, 1948.

\bibitem{Kayser2009Spike}
C.~Kayser, M.~A. Montemurro, N.~K. Logothetis, and S.~Panzeri, ``Spike-phase
  coding boosts and stabilizes information carried by spatial and temporal
  spike patterns.'' \emph{Neuron}, vol.~61, no.~4, pp. 597--608, 2009.

\bibitem{Crumiller2013The}
M.~Crumiller, B.~Knight, and E.~Kaplan, ``The measurement of information
  transmitted by a neural population: Promises and challenges,''
  \emph{Entropy}, vol.~15, no.~9, pp. 3507--3527, 2013.

\bibitem{Nakamura2000Acoustical}
S.~Nakamura, ``Acoustical sound database in real environments for sound scene
  understanding and hands-free speech recognition,'' in \emph{Proc.
  International Conference on Language Resources and Evaluation}, 2000.

\bibitem{G2006The}
R.~Gütig and H.~Sompolinsky, ``The tempotron: a neuron that learns spike
  timing-based decisions.'' \emph{Nature Neuroscience}, vol.~9, no.~3, pp.
  420--428, 2006.

\bibitem{Pons2017Timbre}
J.~Pons, O.~Slizovskaia, R.~Gong, E.~G{\'o}mez, and X.~Serra, ``Timbre analysis
  of music audio signals with convolutional neural networks,'' in \emph{2017
  25th European Signal Processing Conference (EUSIPCO)}.\hskip 1em plus 0.5em
  minus 0.4em\relax IEEE, 2017, pp. 2744--2748.

\bibitem{Bittner2014MedleyDB}
R.~Bittner, J.~Salamon, M.~Tierney, M.~Mauch, C.~Cannam, and J.~Bello,
  ``Medleydb: A multitrack dataset for annotation-intensive mir research,'' in
  \emph{International Society for Music Information Retrieval Conference},
  2014.

\bibitem{cramer2020heidelberg}
B.~Cramer, Y.~Stradmann, J.~Schemmel, and F.~Zenke, ``The heidelberg spiking
  data sets for the systematic evaluation of spiking neural networks,''
  \emph{IEEE Transactions on Neural Networks and Learning Systems}, 2020.

\end{thebibliography}

\end{document}